# Legilimens: Performant Video Analytics on the System-on-Chip Edge

Murali Ramanujam, Yinwei Dai, Kyle Jamieson, Ravi Netravali
*Princeton University*

Continually retraining models has emerged as a primary technique to enable high-accuracy video analytics on edge devices. Yet, existing systems employ such adaptation by relying on the spare compute resources that traditional (memory-constrained) edge servers afford. In contrast, mobile edge devices such as drones and dashcams offer a fundamentally different resource profile: weak(er) compute with abundant unified memory pools. We present Legilimens, a continuous learning system for the mobile edge's System-on-Chip GPUs. Our driving insight is that visually distinct scenes that require retraining exhibit substantial overlap in model embeddings; if captured into a base model on device memory, specializing to each new scene can become lightweight, requiring very few samples. To practically realize this approach, Legilimens presents new, compute-efficient techniques to (1) select high-utility data samples for retraining specialized models, (2) update the base model without complete retraining, and (3) time-share compute resources between retraining and live inference for maximal accuracy. Across diverse workloads, Legilimens lowers retraining costs by 2.8-10× compared to existing systems, resulting in 18-45% higher accuracies.

## 1 Introduction

Machine Learning (ML)-based processing of live video data is the cornerstone of many diverse applications, ranging from autonomous vehicles and traffic coordination to business analytics and disaster relief [22, 26, 33, 88]. Recent years have seen a steady push to shift these video analytics deployments to the network edge, with the primary drivers revolving around enhanced data privacy, tolerance of network fluctuations or outages, and lower response latency [19, 21, 50].

Numerous techniques have been developed to address the innate restrictions that the edge brings [19, 21, 50, 53, 78], i.e., limited resources relative to datacenter servers, and thus, a need to use compressed or specialized models with limited capacity and robustness to data drift. Most notably, existing edge analytics systems aim to continually adapt models to cater to current scene dynamics, all the while supporting live inference for high accuracy and low latency results. To do so, the primary techniques they leverage involve either (1) periodic model retraining at the edge using recent frames [16, 44, 57], or (2) maintaining and swapping among a zoo of historical models that cater to different scene characteristics [43, 72].

Unfortunately, we find that these techniques are fundamentally ill-suited to support the next generation of edge deployments that run analytics directly on data sources themselves [21, 50, 53], e.g., drones, smartphones, or dashcams at the 'mobile edge' that are now routinely provisioned with ML-capable hardware [39, 73, 89]. Indeed, our results show that the accuracy existing systems achieve for a given model decreases by 54-60% when shifting from a traditional edge server to the mobile edge (§2.2). The main issue is that these techniques rely on using the spare compute resources (for periodic retraining or maintaining a model zoo) that are typically present on edge servers which are memory-constrained. In contrast, owing to power and space constraints, mobile edge devices rely on System-on-Chip (SoC) hardware with *weak compute resources but (abundant) unified memory between CPUs and GPUs* (Figure 1 and §2.1). The effect is 71-84% of time being spent on taming drift rather than live inference.

In this paper, we ask whether edge video analytics systems can manage data drift by leaning on abundant memory rather than scarce compute. Our answer is rooted in a key empirical observation (§3): although the raw visual appearance of video streams continues to drift over time, a model's internal representation of scenes – its embedding space – remains remarkably stable. Diving deeper, visually distinct scenes (that warrant retraining) often map to nearby regions in feature space, with a median (75th percentile) cosine similarity of 0.65 (0.78) across scenes in our dataset. This is because convolutional neural networks are explicitly trained to abstract away surface-level variation – e.g., lighting, textures, minor perspective shifts – in favor of structural cues that generalize, such as object contours, spatial layout, and scene geometry [48, 69, 85]. This representational convergence presents an opportunity: if we can accumulate *only* the shared structure across scenes into a model, then retraining for each new scene can be reduced to minor (and thus, computationally lightweight) specialization.

We present **Legilimens**, a continuous learning system for the mobile edge that embodies this idea by drawing on principles from Metalearning [28, 38, 58]. Legilimens tames drift by storing two copies of the analytics model in shared device memory, (1) a *base model* that encapsulates the shared structure across scenes over time, and (2) a *specialized model* that targets the latest scene dynamics. As scene drift unfolds and inference accuracy degrades, Legilimens periodically refines its base model to consider the latest frames, and uses it as a starting point to *rapidly* (i.e., with few examples) train a new specialized model for live inference.

To realize this approach without overburdening the mobile edge's restricted compute resources, Legilimens employs several new techniques. First, Legilimens incorporates an embedding-based, online sampling algorithm that uses only inference-time activations (i.e., no additional compute) to select a small, but high-utility set of recent data samples for few-shot retraining of specialized models. Second, to cope with the mobile edge's lack of space-sharing GPU primitives (e.g., MIG, MPS), Legilimens introduces fine-grained scheduling strategies that capitalize on the rapid accuracy boost that each Metalearning training epoch brings to reactively switch between inference and retraining for maximized accuracy. Lastly, we show in Legilimens how intrinsic characteristics of video scenes and tasks enable high-fidelity base model updates via only (cheap) interpolation from the latest round of specialized model training, rather than complete retraining with (costly) back propagation.

We evaluated Legilimens on 50-hours of video from mobile edge devices, multiple vision tasks (classification and detection), and different physical mobile edge hardware (drone and dashcam SoC GPUs). Compared to state-of-the-art baselines (Ekya [16], RECL [43]), Legilimens improves median accuracy by 18.1-45.1%. These wins stem from Legilimens reducing retraining costs by 2.8-10×.

## 2 Motivation

Video analytics use cases on mobile edge devices like drones and dashcams have risen, with examples such as autonomous navigation [33], disaster recovery [22], search-and-rescue operations [26], and public safety monitoring [16]. This section studies the unique properties of these devices, and how they are ill-served by solutions aimed at traditional edge servers.

### 2.1 Distinguishing Characteristics of the Mobile Edge

Mobile edge devices embed system-on-chip (SoC) hardware that integrate CPU, GPU, and memory onto a single die to meet strict size, power, and thermal constraints [39, 73, 89]. A central distinction between SoC deployments and the discrete GPUs used in prior edge analytics work [16, 43, 44, 57, 72] lies in the nature of the bottleneck: *SoCs are compute-bound, not memory-bound*. This inversion arises from architectural tradeoffs. To stay within tight power envelopes, SoCs prioritize low-power memory access and integration efficiency over raw compute throughput. While DRAM capacity scales relatively easily, increasing compute density greatly raises power and thermal budgets, making it the dominant constraint on these platforms. Figure 1 illustrates this delta, with SoCs often having over 10× less compute capacity than traditional discrete GPUs, but relatively large memory pools.

This asymmetry has two critical implications for system design. First, *model selection is based on compute restrictions*: inference saturates available compute well before exhausting memory, so model selection is dictated by GPU throughput, not memory footprint. Second, *compute-heavy orchestration is impractical*, with lack of support for GPU features such as Multi-Process Service (MPS) or Multi-Instance GPU (MIG) that require dedicated hardware components such as thread schedulers, SM partitioning logic, and fast context-switching units [61, 62]. These components are absent on SoCs due to area and power constraints, preventing concurrent kernel execution and isolated execution streams. As a result, all GPU workloads must serialize through a single queue.

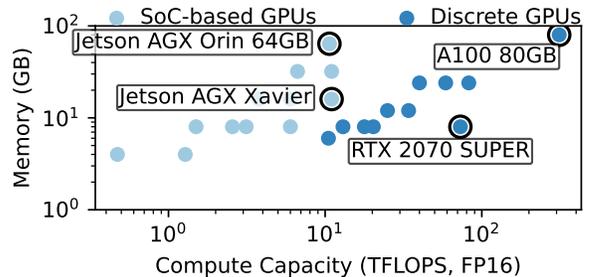

Figure 1: Provisioned memory vs. normalized compute (FP16, Tensor Cores where available) across SoC-based and discrete GPUs. SoC platforms diverge sharply — exposing disproportionately more memory relative to compute [2–11].

### 2.2 The Problem: Handling Drift

Scene drift from changing object classes, lighting, spatial layout, and other visual factors is a persistent challenge in real-world video analytics [16, 43, 72]. To evaluate the efficacy of existing drift solutions on mobile edge devices, we ran experiments on both a Jetson Nano GPU [8] (representing the mobile SoC-based edge), and a full-capacity V100 GPU [12] (representing a traditional edge server). Additionally, we consider an edge server incorporating a V100 with compute artificially capped to match the Nano's capacity; this showcases limitations around lack of scheduling primitives. Results use a ResNet18 model that can run in real time on the mobile edge GPU, and §5.1 details our methodology.

**Approach 1 — High retraining cost from fine-tuning the most recent model.** Systems in this category periodically retrain the latest model used for inference with fresh data samples. We study this approach using Ekya [16] on all three platforms. However, due to the lack of scheduling primitives like MPS on the SoC GPU, we serialize Ekya's training and inference during evaluation, and (favorably) use an oracle to determine the optimal split under time-based scheduling.

As shown in Figure 2, Ekya performs well on the full-capacity V100, but degrades sharply under compute constraints: median accuracy drops by 22% and 60% when shifting to the capped server and Jetson Nano. Figure 3 shows why: as resources shrink, retraining consumes a growing share of GPU time. On the Nano, it dominates—84% of time at the median—leaving inference starved and accuracy degraded.

**Approach 2 — Background overheads from model reuse.** These systems reduce retraining cost by maintaining a zoo of pre-trained models, and reusing the best (i.e., semantically closest) candidate for each scene. The chosen model is then fine-tuned on recent data samples. We study this approach via RECL [43], which uses a gating network to select models based on scene features and retrains them as needed. As above, we serialize its execution on the Jetson Nano and mask storage and selection overheads to isolate the reuse strategy.

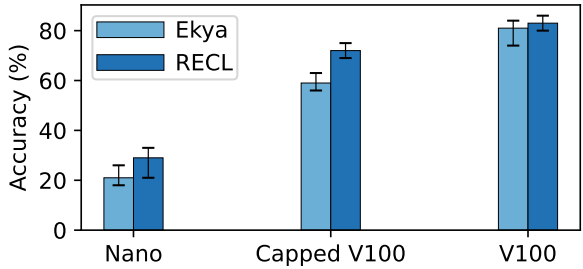

Figure 2: Performance of Ekya and RECL across three hardware settings. Bars denote the median video across our dataset, and error bars span the 25-75th percentiles.

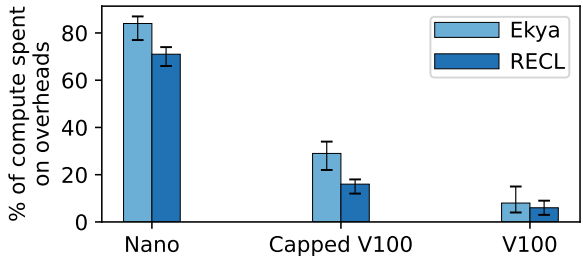

Figure 3: Overheads of retraining (Ekya) and model zoo maintenance (RECL) consume a growing share of compute as the available compute shrinks. Bars denote the median video across our dataset, and error bars span the 25-75th percentiles.

Figure 2 shows that RECL performs well on both server-class GPUs, but accuracy drops sharply (by 54% at the median) on the Nano. To understand why, we examine the hidden cost of managing the model zoo. Each time a new model is added, RECL must update its gating network by evaluating the new model on all prior datapoints and evaluating existing models on new data samples. As shown in Figure 3, this maintenance overhead quickly overwhelms the mobile edge, accounting for 71% of median compute time.

**Takeaway.** Our results highlight that existing approaches stress compute resources to tame drift, making them a poor fit for SoC-based edge devices that end up with few cycles for live inference (and thus, unacceptably low accuracies). Instead, we require new drift strategies that lean into the more abundant memory that these devices afford, while ensuring that compute requirements (both amount and primitives) do not exceed what the mobile edge can support.

## 3 Overview of Legilimens

The key idea driving Legilimens's ability to lean into device memory rather than extra compute is that evolving scenes actually exhibit strong stability in their embeddings, i.e., how a model interprets them. Moreover, starting from that shared representation to specialize to each new scene can drastically reduce the amount of retraining (and data samples) necessary for adaptation. In what follows, we break down this idea from observation to potential benefits.

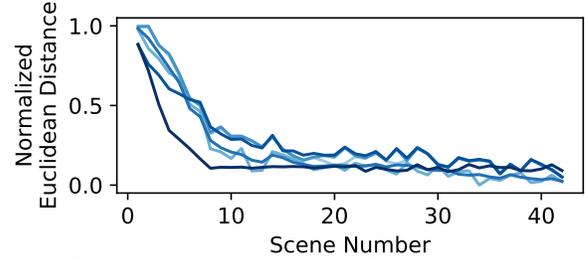

Figure 4: Normalized Euclidean distance between a new scene and the aggregate of all prior scenes, computed from penultimate-layer embeddings of all objects. Results use ResNet18 classification and five representative videos.

### 3.1 Understanding Drift Through Scene Embeddings

**Scene embeddings converge over time.** To study this, we divide each video into 3-minute *scenes*, matching the granularity of retraining we observe that ResNet18 needs to sustain accuracy on our video dataset. For each scene, we compute a centroid vector from the penultimate-layer embeddings (across all detected objects). Figure 4 shows how the distance between each new centroid and the aggregate centroids of prior scenes steadily declines before plateauing. This indicates that as the model sees more scenes, its internal representations begin to saturate. Put differently, while video content continues to drift, the model's interpretation of that content increasingly overlaps with past scenes.

**Why do embeddings converge despite visual drift?** To understand this, we assess each pair of adjacent scenes in our dataset in two ways: (1) comparing embeddings via cosine similarity (as above), and (2) computing pixel-level visual similarity (SSIM) [76] between frames at the scene boundaries. Figure 5 shows that even when SSIM is low (i.e., scenes are visually distinct), model embeddings are often quite similar; Figure 6 shows example screenshots highlighting this property for different lighting, perspectives, weather, and perspective. The reason is that vision processing models are explicitly trained to prioritize (generalizable) structural cues such as object contours/geometry, layouts, and motion corridors rather than surface-level variations [48, 69, 85].

**Embedding convergence can reduce retraining cost compared to prior approaches.** If many distinct scenes map to similar regions in the model's embedding space, then the specialized models adapted to these scenes should also remain close in embedding space. Consequently, the aforementioned shared structural cues (which we call a *base model*) could serve as an effective starting point for all scenes, requiring only lightweight retraining with few samples. To quantify this, we specialize a ResNet18 [34] model for each scene in our dataset in sequence. For each new scene $s_t$, we measure the Euclidean distance in embedding space between its trained model and two initialization points: (1) the specialized model from the prior scene $s_{t-1}$ (as used in Ekya), and (2) the centroid of all specialized models trained for earlier scenes $\{s_1, ..., s_{t-1}\}$ which approximates the base model approach. This metric provides a coarse measure of how much models differ when trained on distinct scenes.

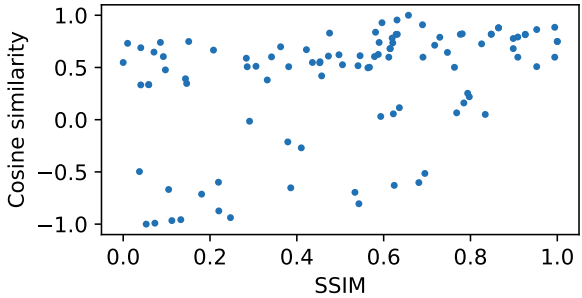

Figure 5: Each point reflects the SSIM and embedding (cosine) similarity for a pair of scenes, averaged over all detected objects. While SSIM varies widely, embeddings remain consistently close. Results shown for ResNet18 classification on 3-minute scenes, uniformly sampled from our dataset.

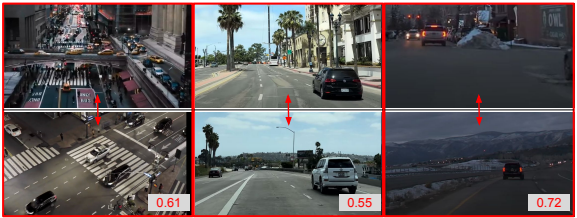

Figure 6: Distinct scenes with high object-level similarity: despite differences in overall imagery, cropped objects are mapped to similar feature space positions by ResNet18. Numbers indicate average cosine similarity between objects, which can range from -1 (opposite) to 1 (identical).

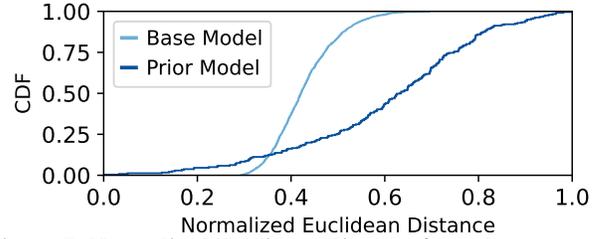

Figure 7: Normalized Euclidean distance from the target specialized model to two initialization points: the base model (Legilimens-style), and prior model (Ekya-style). The base model is closer for a large proportion of the cases, suggesting potentially lower adaptation cost.

Figure 7 shows that the base model is consistently closer to the next specialized model than the previous one is, motivating a path to reduced retraining costs. The core intuition for this behavior is that the prior specialized models (Ekya-style) tend to overfit to prior scene-specific idiosyncrasies — such as transient lighting conditions, rare object classes, or particular spatial layouts — which must be unlearned before the model can specialized to a new scene. In contrast, starting from a base model that preserves general structure requires smaller weight updates, making retraining faster.

### 3.2 Legilimens's Approach

Building on the findings in §3.1, Legilimens maintains *two lightweight models* with the same architecture at runtime: (1) a *base model*, which accumulates shared structure across scenes, and (2) a *specialized model*, adapted for inference on the current scene. When a new scene begins, the system does not train from scratch or reuse the last specialized model. Instead, it initializes the specialized model from the base and fine-tunes only the task-specific aspects. This reduces compute while still enabling accurate adaptation. Moreover, the base model is periodically updated to reflect what was learned during the fine tuning for each specialized model such that it becomes a stronger starting point for future scenes.

Unlike RECL, which maintains a separate model per scene with high compute overheads, Legilimens uses just one persistent base. And unlike Ekya, which re-trains the same model repeatedly without leveraging shared structure, Legilimens's base model acts as a reusable foundation, learned once and refined gradually. Crucially, Legilimens's approach trades off larger memory to maintain two models for lower compute costs when retraining for drift, thereby catering to the unique resource profiles of SoC GPUs.

**Visualizing base model convergence.** To analyze the evolution of the base model over time (and thus, the efficacy of Legilimens's approach), its internal feature representations are examined using a ResNet18 classifier on an examplar 90-minute drone video stream from our dataset. Specifically, as above, we represent a model's view of each scene by extracting all objects and using the centroid of the model's penultimate-layer embeddings [56, 64]; we do this with each specialized model and base model version over time. Comparing these centroids provides an interpretable measure of the base model's alignment with scene distribution over time.

As shown in Figure 8, the base steadily drifts toward the center of the specialized distribution, accumulating cross-scene structure and becoming a stronger initializer for future scenes. This mirrors the convergence trend in §3.1: as the base gets closer, each new scene requires smaller updates, reducing retraining cost over time.

**Connection to Metalearning.** Legilimens's design is a natural instance of metalearning—training models [28, 38, 58] not just to perform well on individual tasks, but to adapt quickly to new tasks using minimal data and compute. In this formulation, each video scene is a task: its content, lighting,

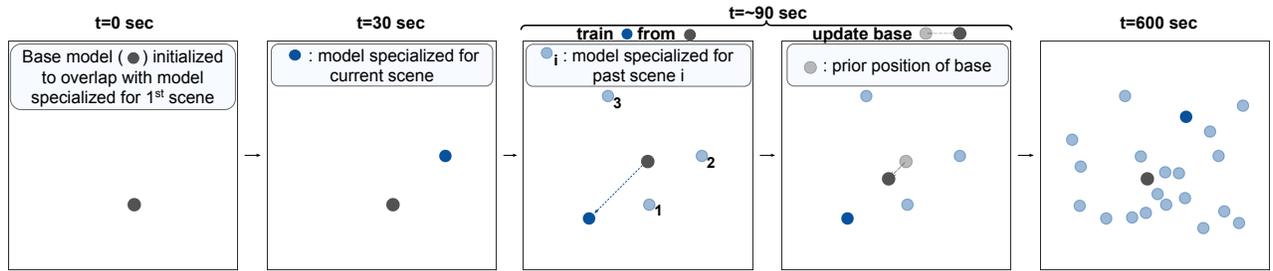

By **t=90 sec**, the ● is already in a "good position", as ●→● (Legilimens-style) is shorter than ●₃→● (training with Ekya-style continual learning).

Figure 8: Evolution of Legilimens's models over time on an example video. Each point represents a model in 2D plane, visualized by averaging ResNet18 penultimate-layer features across all detected objects in a scene. Features are projected into 2D using PCA [42]; axes correspond to the top two principal components. While absolute positions are not meaningful (axes are unlabeled), distances between points reflect changes in the model's internal representation.

layout, and object composition differ from prior scenes, often subtly, sometimes substantially. This maps directly to the goals of metalearning algorithms like MAML [28], which optimize for fast adaptation by learning a generalizable initialization. In Legilimens, that initialization lives in memory as the base model, and each per-scene retraining episode acts as an inner-loop adaptation step. Moreover, the results above align with prior meta-learning literature [29, 59, 65], which shows that effective initializations naturally converge toward representations that minimize adaptation distance across task distributions. Indeed, we empirically observe that after convergence, the base model lies within 4.7% (based on mean Euclidean distance between embeddings) of the average specialized model across scenes, suggesting that it serves as a representative initializer.

### 3.3 Challenges

Although metalearning enables rapid adaptation, realizing it efficiently on SoC-class devices involves practical challenges that span the end-to-end pipeline.

**Challenge 1: Sensitivity to Training Data Selection.** Metalearning enables rapid adaptation with minimal gradient steps and few labeled datapoints. However, this few-shot setting is fragile: success depends on selecting datapoints that precisely reflect the shift between the new scene and the base model's prior knowledge which is difficult to guarantee in practice. For instance, uniform random sampling from the new scene leads to high variance in outcomes; simply changing the random seed across 100 trials causes accuracy to vary by up to 38%. Oversampling mitigates variance but negates the efficiency benefits of few-shot learning. Thus, Legilimens must select datapoints that are both diverse and informative, capturing novelty without redundancy. Yet precise selection can be costly since utility is based on whether a base model already captures the semantics of a given sample, i.e., we must determine how confident the base model is for each new sample, which would involve impractical forward passes.

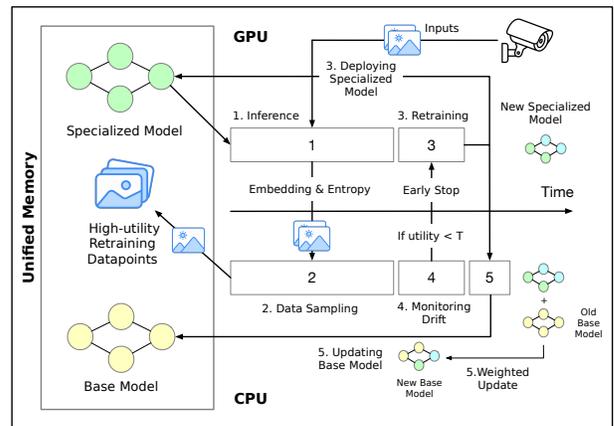

Figure 9: Overview of Legilimens's end-to-end workflow.

**Challenge 2: Lack of scheduling primitives.** As discussed in §2.1, SoCs only support coarse-grained time sharing of GPU resources, forcing serialization. Thus, even short bursts of retraining will interfere with (i.e., pause) live inference, leading to latency spikes and lower throughput, or lower accuracy as shown in Figure 2. Thus, to remain responsive, a system must not only reduce the duration of retraining but also swiftly capitalize on instances when preempting retraining in favor of inference is advantageous.

**Challenge 3: Efficient and Robust Base Model Updates.** Meta-learning systems must continually update a shared base model to accumulate structure across tasks. However, standard approaches like MAML [29] rely on higher-order gradient computations that increase training time by 3–5× [59], making them infeasible on resource-constrained SoCs. Updating the base must therefore be lightweight, yet effective enough to retain useful generalization across scenes. In addition, updates must remain stable under bursty or short-lived scene changes, without overfitting to transient scenes.

## 4 Design

Figure 9 shows Legilimens's operation at a high level. As video streams from the mobile camera, Legilimens, operating on-device (drone or dashcam) serves predictions using a lightweight, scene-specialized model on the GPU. Users register queries with an on-device agent, specifying the model architecture (e.g., ResNet18), objects of interest, and a task (e.g., classification or detection).

To remain accurate over time, Legilimens must respond to scene drift (§2). It does so by periodically pausing inference ❶ — every 30 seconds in our setup[1] — and retraining ❸ the specialized model. To keep this retraining lightweight, Legilimens uses Metalearning: it starts retraining from a reusable base model that encodes prior structure and adapts it to the new scene using a small number of datapoints.

Training examples are selected in advance on the CPU by an online sampler ❷ that runs during inference and filters for diverse, high-utility datapoints (§4.1). Since manual labeling is infeasible, as with prior approaches [16, 43], Legilimens uses a golden teacher model (e.g., ResNeXt101) to label this set just before retraining (§2.2).

Retraining occurs on the GPU, while a drift monitor ❹ continues in parallel on the CPU. Legilimens monitors both model progress and scene dynamics during training, and may halt early if accuracy gains plateau or scene conditions change (§4.2). Once retraining completes (or exits early), inference resumes immediately with the updated model on the GPU ❸. The base model's weights are then updated ❺ toward the new weights using a lightweight update rule (§4.3) on the CPU, improving generality over time.

### 4.1 Selecting Training Datapoints

Retraining on the mobile edge must be fast, accurate, and minimally disruptive to inference. To meet these constraints, Legilimens retrains on a small but carefully selected set of datapoints—large enough to capture scene variation, yet compact enough to remain tractable on limited compute. Figure 10 quantifies this tradeoff. As expected [71], accuracy improves with more datapoints, but plateaus quickly—while retraining time continues to grow linearly. For example, increasing the training set from 200 to 500 datapoints improves accuracy by less than 3%, but more than doubles compute cost. These results highlight a central challenge: retraining on too many datapoints incurs high cost, while uninformed down-sampling can miss out on accuracy wins.

Legilimens addresses this challenge with a structured two-stage pipeline: it first removes redundant datapoints using object-level embeddings, then prioritizes the most informative among them based on model uncertainty. This process runs in parallel with inference and allows the system to construct a concise training set with minimal additional overhead.

---
[1]Periodic retraining is common in systems like Ekya and RECL. In general, drift detection could be used to trigger updates; we treat this as a policy decision orthogonal to our design.

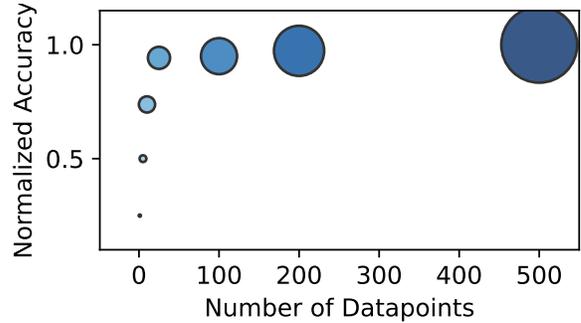

Figure 10: Post-retraining accuracy and relative retraining time (circle area) as a function of number of training datapoints for a ResNet18-based classifier (averaged over scenes from a 1 hour drone stream in our dataset). Accuracy improves logarithmically, while training time grows linearly, highlighting diminishing returns from larger updates.

**Eliminating redundant datapoints.** From Figure 10, training on too many datapoints is wasteful and slows down adaptation, especially when compute is scarce. To effectively eliminate redundant samples (objects re-appearing across successive frames, or multiple instances of the same class within a frame), prior systems [15, 40, 82] have explored visual similarity heuristics—such as frame differencing or background subtraction. However, these methods are brittle in mobile settings where both foreground and background change frequently. More advanced alternatives, like optical flow [35] or segmentation [46], offer better accuracy by capturing fine-grained visual variation, but are prohibitively compute-intensive for SoC-based devices. For example, optical flow can nearly double the end-to-end retraining time, while simpler heuristics can incur accuracy drops of up to 22%.

A key observation here is that the model's own internal representations offer a natural lens for assessing redundancy. As objects are processed during inference, the specialized model produces embeddings in its intermediate layers that encode how it semantically interprets each object — capturing cues like shape, edges, structure and color. If these embeddings are similar, the datapoints are likely redundant from the model's perspective. This presents the opportunity to filter training candidates directly in embedding space, yielding a principled, model-aligned approach to sampling diversity. However, this raises a challenge: retraining starts from the base model, not the current specialized one. To ensure alignment between training data and the base model's perspective, a natural approach would be to recompute embeddings using the base model before retraining. But this incurs a second forward pass per datapoint — adding 1.3× overhead to the retraining.

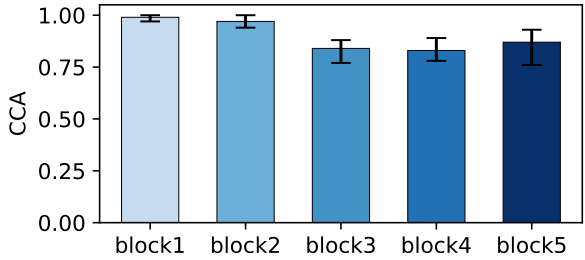

Figure 11: CCA scores between "blocks" of base and specialized models. Each "block" refers to a group of convolutional layers, with block 1 capturing early features and block 5 closer to the output. Values close to 1 indicate similar representations. Results are for classification (ResNet18) across our dataset. Bars denote the median, error bars show 25th–75th percentiles.

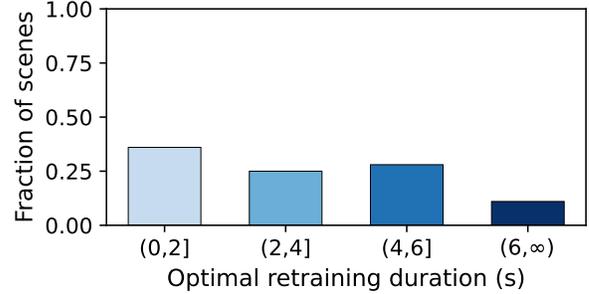

Figure 12: Results list a PDF (binned by 2 seconds) of optimal retraining durations to achieve maximal post-retraining accuracy.

Legilimens avoids this cost by leveraging a structural property of meta-learned models: the early layers of the specialized and base models remain remarkably stable across scenes. As shown in Figure 11, Canonical Correlation Analysis (CCA)[18] reveals strong similarity between the corresponding early layers. These layers capture stable, low-level features—such as object edges, contours, and textures—that generalize across scenes and persist through adaptation.

This observation enables Legilimens to reuse embeddings computed during inference by the specialized model. Cached in shared memory, these embeddings offer a semantically rich and low-cost representation of each datapoint — sidestepping the need for recomputation from the base model. Legilimens leverages an online variant of spatially correlated poisson sampling (SCPS) [32] — setting the inclusion probability of a datapoint to be inversely proportional to the distance to other datapoints helps curate a sparse but representative set of datapoints. As new embeddings arrive, each is compared to a buffer of recently selected points and retained only if sufficiently dissimilar. The SCPS *sampling threshold* is adapted dynamically based on the distribution of pairwise distances: when embeddings are tightly clustered (indicating low scene diversity), fewer points are accepted; in more variable scenes, the system accepts a broader set. The deduplication process is lightweight enough to run continuously on the CPU—maintaining a sparse, diverse training set in real time without blocking GPU-based inference.

**Prioritizing informative datapoints.** After deduplication, Legilimens holds a small, diverse buffer of candidate datapoints for retraining. However, not all datapoints contribute equally to model improvement; those that align with the current base model's predictions yield little new gradient signal during training. To select those with the highest utility, Legilimens applies an active learning heuristic with the base model — entropy over output probabilities — to approximate model uncertainty.

This approach is grounded in the observation that the *base model already encodes general structure shared across scenes* (§3). Training a specialized model requires focusing on the remaining information — the scene-specific delta. High-entropy predictions indicate areas where the base model is uncertain and least confident, making them the most informative.

Legilimens applies this filtering across task types (e.g., entropy for classification, confidence thresholds for detection), selecting the top 5% for labeling via the golden model. These labeled examples form the training set for the specialized model. Because this filtering step operates over a small, deduplicated pool, it adds minimal overhead and can therefore briefly preempt inference on the GPU just prior to retraining.

### 4.2 Fast, Inference-Aware Retraining

After selecting a small, high-utility training set, Legilimens initiates retraining of a new scene-specialized model. Although retraining is lightweight, serialized GPU execution on SoCs means inference must pause during training. This introduces a core tradeoff: longer retraining can improve model quality, but extended pauses reduce the number of frames processed in real time and increase the risk of drift. As the scene continues to evolve, the model may end up adapting to stale datapoints, reducing its utility once retraining completes. Importantly, this tradeoff varies by scene. As shown in Figure 12, the optimal stopping point differs significantly across scenes, underscoring the need for an adaptive policy.

**Opportunistic Early Stopping.** Metalearning offers a distinctive property: rapid accuracy gains when compared against traditional continual learning, as shown in Figure 13. This behavior stems from the fact that training begins from a base model already tuned to the structure of past scenes, requiring only modest fine-tuning to adapt to the new task.

Legilimens leverages this property to implement *fine-grained, intra-training control*. After each epoch, Legilimens evaluates whether to continue or halt training based on a stopping score that balances two competing forces: the marginal accuracy gain from additional training and the degree of scene drift observed during the retraining window. If further adaptation offers diminishing returns while drift increases, retraining is terminated early, and inference resumes with the current model checkpoint.

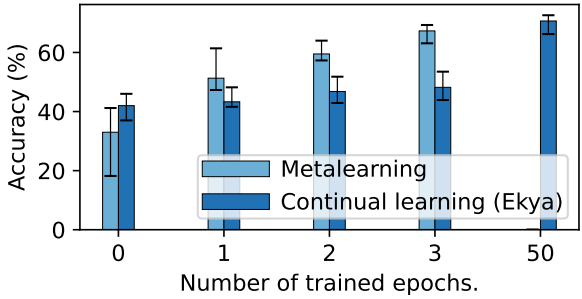

Figure 13: Accuracy ramp-up with metalearning is significantly faster than with traditional continual learning, which repeatedly trains a single specialized model. Bars show classification accuracy (ResNet18) on the median video, with error bars spanning the 25th–75th percentiles. Meta-learning converges well before epoch 50; its bar is thus omitted.

Legilimens uses opportunistic early stopping to make time-sharing tractable — identifying the point where further retraining yields diminishing returns and resuming inference promptly. Formally, the stopping score at epoch $t$ is:

$$S_t = w_1 \cdot \frac{\Delta A_t}{\Delta A_{\max}} - w_2 \cdot \frac{D_t}{D_{\max}}$$

- $\Delta A_t$: marginal accuracy improvement from epoch $t-1$ to $t$
- $D_t$: feature-level drift computed from incoming scene
- $w_1, w_2$ balance training benefit versus inference urgency

Retraining halts when $S_t \leq \tau$, a tunable threshold (0.1 in our setup). This avoids over-committing to training while blocking inference — necessary on the time-shared GPU. Legilimens's early stopping policy adapts to scene dynamics. When the incoming scene closely resembles prior environments, the system halts retraining quickly — since little adaptation is needed. Conversely, when scenes are highly dynamic or fast-changing, prolonged retraining yields diminishing returns, as the specialized model quickly becomes stale. In between these extremes lies a sweet spot, where modest retraining can significantly improve accuracy without blocking inference. As we show in §5.3, this enables Legilimens to selectively allocate retraining effort where it matters most — recovering inference throughput without degrading accuracy.

### 4.3 Updating the Base Model Efficiently.

Once retraining completes, Legilimens switches the GPU back to inference, deploying the newly trained specialized model for real-time predictions. In parallel, it updates the shared base model on the CPU using a lightweight meta-learning step based on Reptile [59], a first-order algorithm that avoids expensive gradient computations. The vanilla Reptile algorithm updates the base model using lightweight interpolation between its current weights and the specialized model's weights — but, as outlined in §5.3, this method is brittle in the face of transient drift; i.e., a momentary variation in the scene could harm the base model convergence. To alleviate this issue, Legilimens augments Reptile to insulate it against such temporary scene drifts using a dynamic exponentially weighted moving average (EWMA [1]) factor. In this technique, the EWMA weight $\epsilon_t$ is dynamically chosen based on *scene similarity*. If the current scene is similar to the previous one (cosine similarity between scene embeddings $s \geq 0.9$), we set $\epsilon_t = 0.3$ to incorporate the new structure more aggressively. If the scene appears dissimilar — indicating a transient event — we use a more conservative $\epsilon_t = 0.05$ to avoid overfitting. Legilimens's Reptile variant is sketched in Appendix.

Both sets of embeddings used in determining *scene similarity* are readily available: the prior scene's is already stored, and the current scene's is generated as inference begins — avoiding any additional compute. The final interpolation update to the base model $\phi$ is computed from the specialized model $\zeta$ as $\phi \leftarrow (1 - \epsilon_t) \cdot \phi + \epsilon_t \cdot \zeta$, smoothly interpolating the base model toward the specialized weights. This drift-sensitive update stabilizes the base model over time while allowing for faster adaptation when true distribution shifts occur. It improves robustness to bursty content (e.g., tunnels, weather, rare events) while allowing timely response to actual context changes (e.g., merging onto a freeway). Importantly, this update runs entirely on the CPU, avoids backpropagation, and requires only a single vector interpolation — ensuring that inference on the GPU is never interrupted. Over time, this allows the base model to shift toward the centroid of the specialized models, providing a stable and increasingly generalizable initialization for future adaptation.

In general, for tasks with large or complex task distributions, first-order methods like Reptile can lead to accuracy losses compared to MAML due to their coarser credit assignment [20, 38]. However, Reptile suffices in our setting, offering nearly identical downstream accuracy (within 0.5% of MAML) while reducing compute overhead by 3-5×.

**Why Reptile Suffices for Legilimens.** Reptile approximates the behavior of gradient-based meta-learning (e.g., MAML) by forgoing higher-order derivatives and instead interpolating the base model toward the final weights of each adapted model. While this can reduce fidelity in general, Legilimens's setting makes Reptile a natural fit:

- **Stable low-level structure.** As shown in §3.1, many scenes share mid-level visual patterns (vehicle contours, signage,

and road boundaries) captured by early layers of the network. As Reptile updates the entire model toward task-specific weights without relying on fine-grained gradient alignment, it efficiently reinforces these shared features, alleviating the need for MAML's added precision.

- **Recurring environments and object distributions.** Mobile video analytics often revisits structurally similar environments — e.g., multiple passes over intersections or highways. In this setting, the ability to accumulate shared structure over time is useful. Reptile naturally benefits from this recurrence by continuously pulling the base toward a central, reusable representation—whereas MAML's per-task optimization offers diminishing returns.

- **Compact models reduce the need for fine-grained updates.** On SoC-class edge devices, models are kept small (e.g., ResNet18) due to resource constraints. In these low-capacity regimes, the fidelity gains from precise, gradient-aligned updates — as performed by MAML — are often negligible. Empirically, we find that Reptile achieves comparable accuracy, making the added complexity of higher-order optimization unnecessary in practice.

## 5 Evaluation

We evaluate Legilimens across real-world video workloads, model architectures, and hardware platforms.

- Legilimens achieves 42.7-45.1% and 18.1-26.8% higher median accuracies than Ekya and RECL via 2.8-10× lower retraining costs; wins are within 7-16% of the optimal.
- Ablation studies show that datapoint selection and inference-aware retraining reduce adaptation time by up to 3×, while Reptile-style base model updates reduce CPU cost by 6× compared to MAML-style updates.
- The base model converges within minutes of deployment and can be reused across videos, reducing future adaptation time by up to 26.9%.

### 5.1 Methodology

**Platforms.** We evaluate Legilimens on two representative hardware platforms that reflect real-world edge deployment settings and prior system baselines. Trends for results shown on one (due to space constraints) hold for the other.

- **Jetson Nano (Drone SoC):** Our experiments use a standalone Jetson Nano[8] (128-core NVIDIA Maxwell GPU, quad-core ARM Cortex-A57 CPU, 4 GB LPDDR4 memory + 4 GB swap), which is widely integrated into real-world aerial systems (e.g., Aurelia's drone add-ons and DJI's M350 RTK) via off-the-shelf carrier boards or internal interfaces. The Jetson Nano serves as the default platform for all results not explicitly attributed to a specific device.

- **Adreno 506 (Dashcam SoC):** To represent a lower-end compute platform compared to the Jetson Nano, we leverage the Qualcomm Snapdragon 450 platform[60], commonly used in dashcam-grade devices such as the Novatek-based K145/K245. Across GPU benchmarks, this platform achieves up to 8× weaker performance than the Jetson Nano [60, 63]. The SDM450 includes an octa-core Cortex-A53 CPU, Adreno 506 GPU, and 4 GB RAM.

**Workloads and Models.** We evaluate two core video analytics tasks: (1) multi-class object classification and (2) object detection. For each task, we select two lightweight models representative of SoC-class deployments, drawn from distinct architecture families. For classification, we use models from the ResNet[34] and MobileNet[67] families (ResNet8, MobileNetV2-lite on the dashcam and ResNet18, MobileNet-V2 on the drone) — each chosen as the largest variant that achieves real-time inference (15+ FPS) on the respective device. Similarly, for detection, we use models from the YOLO [66] and SSD-MobileNet [52] families (YOLOv2-tiny, SSD-MobileNetV2-lite for the dashcam; and YOLOv4-tiny [17], SSD-MobileNetV2 for the drone). We employ ResNeXt101 [81] for classification and YOLOv4 for detection as golden teacher models for ground-truth labeling. While our experiments focus on classification and detection, Legilimens's underlying approach is model-agnostic and extends to other tasks such as segmentation or tracking.

**Dataset.** Our dataset spans over 50 hours of drone and dashcam video footage sourced from public YouTube videos. The dataset covers diverse scenes including urban, suburban, off-road, and natural landscapes from around the world.

**Baselines.** We compare Legilimens with the following:

- Ekya [16]: periodic retraining with a single specialized model. The original design assumes support for MPS or fine-grained scheduling to overlap training and inference. Since the Nano lacks these primitives, we serialize retraining and inference. To approximate the original behavior, we allow inference to "catch up" after retraining (replaying missed frames without skipping) thus maintaining full video coverage while respecting real-time compute limits.

- RECL [43]: model zoo-based reuse with periodic retraining. The original RECL system assumes cloud-scale compute and storage. For SoC evaluation, we provide generous off-device memory for storing the model zoo and remove all selection overhead by assuming an oracle gating network that always picks the best expert. We similarly serialize retraining and inference to account for the lack of MPS.

- Oracle Upper Bound: an idealized system that selects the best possible specialized model for each scene from a large

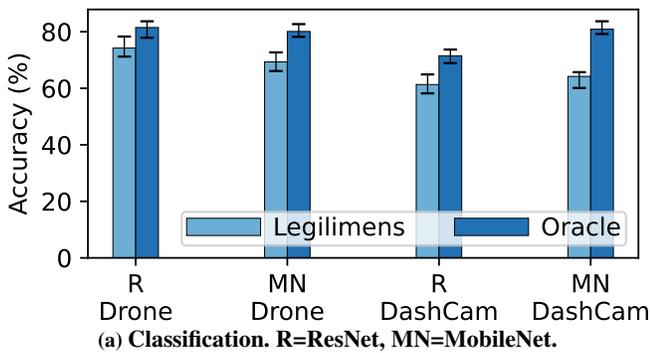
(a) Classification. R=ResNet, MN=MobileNet.

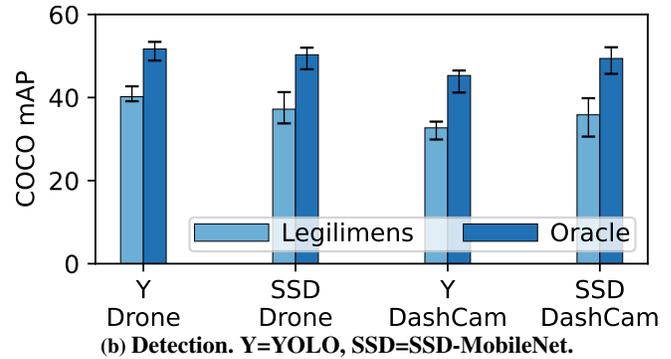
(b) Detection. Y=YOLO, SSD=SSD-MobileNet.

Figure 14: Results compare Legilimens to an Oracle upper bound across all videos. Bars show medians; error bars span the 25–75th percentiles. For each model family (ResNet, YOLO, etc.), the largest variant that supports real-time inference (over 15 FPS) is selected per device, as detailed in § 5.1. For example, "ResNet–Drone" denotes ResNet18 on the Jetson Nano, while "ResNet–Dashcam" uses ResNet8 on the Adreno 506.

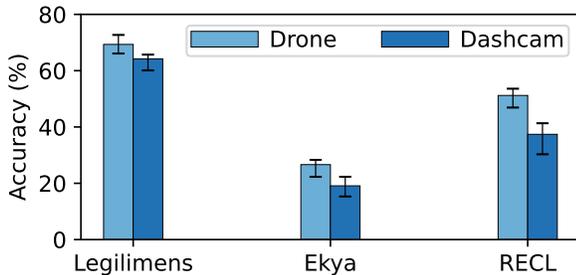

Figure 15: Bars denote median overall classification accuracy (MobileNetV2) per video, error bars show 25th-75th percentile.

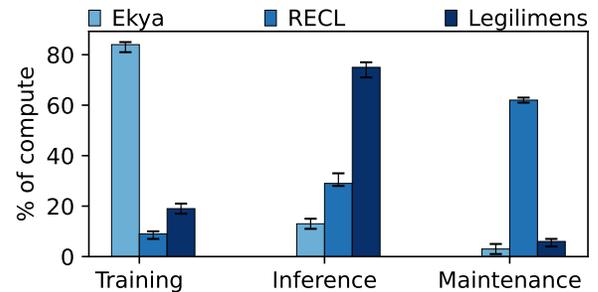

Figure 16: Compute corresponding to Figure 15 (drone).

offline-trained pool, assuming unlimited compute and perfect future knowledge. This represents a theoretical (but impractical) upper bound on scene-specific accuracy.

The Ekya and RECL configurations are intentionally optimistic — designed to show a favorable analysis of each system under best-case adaptation to the SoC setting.

**Metrics.** As with prior work[16, 43, 72], we compare the edge inference results with labels generated offline using expensive golden models. We extract per-frame classification accuracy (fraction of correct predictions) and detection mAP (mean Average Precision) scores and average them over across all frames for overall results.

**Implementation.** Legilimens's core components are written in 3.1k lines of Python code. On the Jetson Nano, we use PyTorch 1.8.1 for training and inference. The system uses NVIDIA's Jetson SDK [23], including JetPack 4.6 (L4T 32.6.1), which bundles CUDA 10.2 and cuDNN 8.2.1. The Snapdragon platform uses ExecuTorch (PyTorch) [55] with Qualcomm Adreno 506 GPU drivers [60]. On both platforms, video frames are processed at 15FPS. CPU-side scheduling, sampling, and monitoring tasks are managed using lightweight multiprocessing primitives (threadpools and shared memory queues) while GPU access is coordinated via shared flags across CPU threads.

### 5.2 Overall Results

As shown in Figure 14, Legilimens achieves median accuracies of 64.2-74.2% for classification and 35.8-40.2 (mAP) for detection, reaching within 7-16% and 11-14 (mAP) of the unachievable oracle. These gains hold across model architectures and platforms, across our dataset indicating robustness to workload characteristics. Importantly, these results demonstrate Legilimens's ability to operate effectively within the compute constraints of SoC-based devices, delivering real-time predictions while continually adapting to evolving video content. Legilimens's performance drops minimally as available compute shrinks from the Drone to the Dashcam—this is due to its low-overhead adaptation pipeline.

Compared against the prior state-of-the-art systems, from Figure 15, Legilimens delivers speedups that are 18.1-45.1% better at the median. Digging deeper, our results show a key trend: as available compute capacity decreases (the Drone is better provisioned than the Dashcam), the performance of the prior state-of-the-art systems shrinks by up to 13.8% compared to the measured degradation experienced by Legilimens (4%). The reason (outlined in Figure 16) is that the compute-intensive retraining (or maintenance) tasks of the prior systems occupy a larger fraction of the available compute, causing to aggressive preemption of the inference tasks, leading to their diminished accuracy. In comparison, Legilimens's retraining (and base model maintenance) tasks are lightweight (retraining costs are lower by 2.8-10×), and lead to a milder loss in accuracy as even at the low compute capacity.

### 5.3 Robustness to Transient and Permanent Drift

We construct synthetic videos by stitching together distinct 30 second video segments (e.g., S1 = sunny highway drone, S2 = snowy city dashcam, S3 = rural drone footage) to evaluate Legilimens's robustness/responsiveness to transient/permanent drift. Figure 17 shows accuracy across 50 such synthetic videos. S1 → S2 → S1 is a transient shift, followed by a permanent transition to S3. The first S1 segment starts with a converged base model.

During the brief shift to S2, all methods experience an expected drop in accuracy — none had prior exposure to that environment. However, key differences emerge when the stream returns to S1. The Vanilla strategy (full update) overfits to the transient S2, degrading accuracy upon returning to S1. In contrast, Static Weights (fixed EWMA, $\epsilon$ = 0.1) resists over-adapting and preserves better performance on S1, but will react slowly to future shifts. Legilimens (dynamic EWMA) strikes a balance — maintaining high accuracy on the return to S1 while avoiding overfitting.

The second half of the video initiates a sustained shift to a new environment, S3. In the first S3 segment, accuracy drops uniformly across all methods. By the second S3, both Vanilla and Legilimens have adapted and recovered performance. Static EWMA also improves, but more slowly — lagging due to its conservative fixed update weight. Legilimens adapts quickly here because it lowers the $\epsilon$ weight factor only during isolated scene shifts, but uses to higher values when the same scene persists — accelerating convergence without prematurely destabilizing the base model.

Digging into the reason for this behavior, Figure 18 quantifies the retraining time during the corresponding scenes. All methods suffer during the first transient burst (S2). However, when returning to the original scene (S1), Vanilla remains slow due to overfitting, while Static Weights recovers partially and Legilimens fully recovers. During the true distribution shift (S3), all methods again initially experience a retraining spike. Yet, by the second and third S3 segments, Legilimens and Vanilla stabilize quickly, whereas Static Weights adapts more slowly. These trends mirror the accuracy patterns seen earlier and highlight how Legilimens balances responsiveness and stability in dynamic environments.

### 5.4 Ablation study

Table 1 quantifies the contribution of each core component in Legilimens's design: (i) datapoint selection, (ii) inference-aware retraining, and (iii) efficient base model updating. Each row disables one component while keeping the others intact. As shown, all three components are critical to performance—removing datapoint selection, inference-aware retraining, or Reptile-style base model updates degrades median accuracy to 39.7%, 47.6%, and 61.5%, respectively.

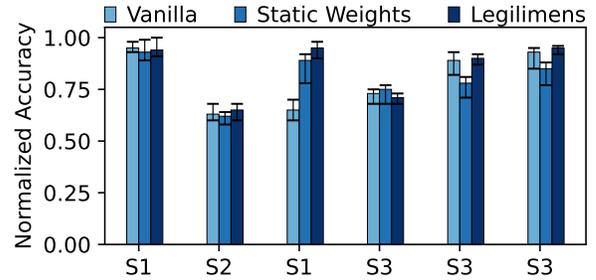

**Figure 17: Normalized accuracy across synthetic scene shifts. Bars show median, errors span 25th–75th percentiles.**

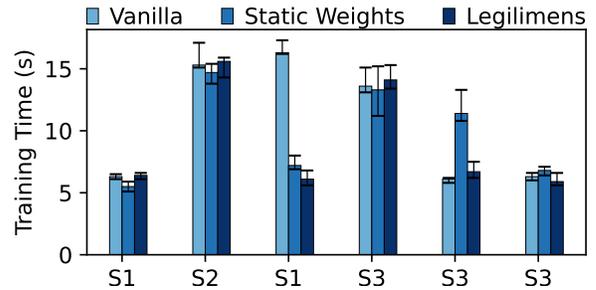

**Figure 18: Retraining times corresponding to Figure 17**

The largest performance drops stem from disabling datapoint selection and inference-aware retraining. On SoC platforms with serialized execution, retraining and inference cannot run concurrently; longer retraining windows directly reduce inference throughput. Legilimens's datapoint selection reduces training set size by 8–14× compared to uniform random sampling, while maintaining comparable accuracy. Similarly, inference-aware retraining adapts to scene dynamics and terminates early when additional training yields diminishing returns—reducing retraining time by up to 3× without hurting final accuracy.

Efficient base model updating is less visible in immediate accuracy as base model updates are pinned to the CPU. Disabling Legilimens's lightweight Reptile-style updates in favor of full MAML-style gradient-based meta-updates increases CPU compute demand by 6× on average. This not only limits how often adaptation can occur but also reduces the compute budget available for the other CPU tasks, including datapoint selection and drift monitoring.

### 5.5 Additional Results

**Base Model Convergence and Reuse** We define base model convergence as the point at which per-scene adaptation becomes compute-efficient—requiring under 5 seconds of training on the Jetson Nano while maintaining high accuracy. We choose 5 seconds as a practical threshold, as it enables frequent retraining (e.g., every 30 seconds) without significantly disrupting inference. This threshold marks the point where the base model has accumulated sufficient cross-scene structure to serve as a robust, general-purpose initializer. From Figure 19 (New Base curve), when beginning with a new base model, Legilimens's base model converges within 260s for the median video.

| Disabled Component | Accuracy (Median, 75%ile) |
|---|---|
| None (full Legilimens) | 69.3% (72.7%) |
| Datapoint selection | 39.7% (43.1%) |
| Inference-aware retraining | 47.6% (52.8%) |
| Efficient base model update | 61.5% (65.4%) |

Table 1: Impact of Legilimens's components. ResNet18 for classification on Jetson Nano.

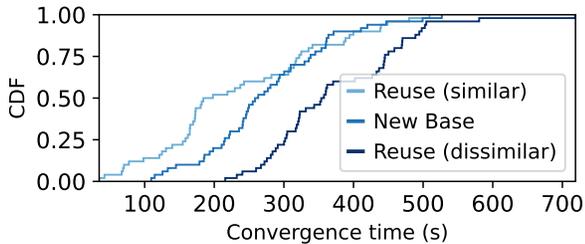

Figure 19: CDF of the time until base model convergence. Evaluated on ResNet18 classification across our dataset.

Figure 19 also shows the impact of reusing a previously converged base model across deployments (as the Reuse (similar) curve). When reused across similar environments (e.g., city to city), convergence is accelerated to 190s (26.9% faster) at the median, requiring fewer adaptation rounds to stabilize. In contrast, reusing a base (Reuse (dissimilar) curve) from a dissimilar environment (e.g., city to rural) slows convergence to 355s, due to structural mismatches in the embedding space. These findings reinforce the importance (and the opportunity) of environmental alignment when transferring base models. Together with our robustness experiments in §5.5, these results suggest that Legilimens strikes an effective balance: it retains past knowledge when environments persist, and resets efficiently when true shifts occur—supporting both fast adaptation and stability under drift.

**Additional results.** The appendix presents results for managing multiple base models and combining Legilimens with RECL. Augmenting Legilimens with multiple bases or a model zoo offers modest accuracy gains at the expense of significant additional compute or memory costs.

## 6 Related Work

**Model optimizations.** Much prior work has focused on optimizing machine learning models. Some techniques compress models by creating light-weight variants of models that maintain accuracy within acceptable margins. Such methods include quantization, which reduces the precision of numerical weights [31, 45, 51]; distillation, which transfers knowledge from larger models to smaller, efficient models [37, 68]; and pruning, which removes redundant model weights or layers to minimize inference cost [30, 54]. Other techniques boost model efficiency by adapting model execution at different granularities [24, 25, 27, 36, 75, 83]. Moreover, compiler-driven approaches [13, 14] optimize model execution at the computational graph or operator level. Legilimens is complementary to these approaches and optimizes sustainable continuous retraining under compute-constrained conditions.

**Other video-analytics systems.** There have been multiple approaches to maintain high inference accuracy while reducing the resource needs, including configuration adaptation [41, 47, 77, 86], DNN feature reuse [40, 82], data filtering [49, 70, 87]. Similarly, Legilimens provides a complementary approach, as Legilimens focuses on efficient retraining. Other approaches rely on continuously retraining the specialized model to fit the change in scenes [16, 44, 57], or augmenting the continual learning process by choosing from previously retrained models [43, 72]. When deployed on a resource-constrained edge, these approaches are suboptimal due to their significant computational demands. Legilimens mitigates the computation cost by leaning on relatively abundant memory on edge devices.

**Resource allocation for DNNs.** Scheduling and provisioning mechanisms target both multi-tenant training clusters and latency-critical inference services. Gandiva [79] introduces domain-aware GPU time-slicing and migration to raise utilization in shared clusters. Building on this, Salus [84] and AntMan [80] expose fine-grained GPU-memory lanes and elastic scaling to dynamically resize a job's footprint, boosting throughput without compromising fairness [80, 84]. GPUPool [74] refines the packing problem by co-scheduling kernels from different jobs to saturate modern accelerators. However, these approaches don't directly generalize to the resource-constrained edge due to hardware limitations, which is the focus of this work.

## 7 Conclusion

Legilimens revisits the problem of continual adaptation for video analytics in resource-constrained edge settings. Unlike prior systems that rely on compute-intensive retraining or large-scale model reuse, Legilimens embraces the inverted resource profile of SoC-class devices—shifting the cost of adaptation from compute to memory. By combining a persistent base model with embedding-guided sampling, inference-aware retraining, and lightweight meta-updates, it delivers accurate, per-scene adaptation in real time, without disrupting inference. Our evaluation shows that Legilimens lowers retraining costs by 2.8-10× compared to existing systems, resulting in 18-45% higher accuracies across diverse workloads.

## A Appendix

We list the pseudocode for Legilimens's base model update.

---
**Algorithm 1** Reptile Meta-Update in Legilimens
---
1: Let $\phi$ be the base model weights
2: **for** epoch $i = 1$ to $T$ **do**
3:     Train specialized model from $\phi$ on current scene $\rightarrow \zeta$
4:     Break if early stopping criterion (see §4.2)
5: **end for**
6: Compute scene similarity $s$ between current and previous scene embeddings (reused from inference).
7: **if** $s > \tau$ **then**     ▷ Scene is similar (persistent)
8:     $\epsilon \leftarrow 0.3$
9: **else**     ▷ Scene is dissimilar (bursty)
10:     $\epsilon \leftarrow 0.05$
11: **end if**
12: Update base: $\phi \leftarrow (1 - \epsilon) \cdot \phi + \epsilon \cdot \zeta$
13: Return updated $\phi, \zeta$

---

### A.1 Additional base models.

We also evaluated whether maintaining more than one base models on device could improve adaptation performance, even within a single video stream. In this setup, two base models are stored in memory, and an oracle selects the better of the two for each scene, based on which yields lower validation loss after adaptation. This configuration improves accuracy by 2.4% on average over the single-base variant, suggesting that multiple bases can help modestly when scene content varies substantially even within a single deployment. However, this result assumes perfect base model selection at runtime; we do not account for the cost or feasibility of implementing such a selection mechanism. Moreover, the memory footprint increases proportionally with the number of bases. At the extreme, with high memory availability and a growing pool of environment-specific experts, this approach starts to resemble RECL's model zoo, which we evaluate next.

**RECL + Legilimens.** We next tested a hybrid approach that combines RECL's model reuse mechanism with Legilimens's metalearning-based adaptation. In this setting, RECL's model zoo is still used, but each selected expert is further adapted using lightweight meta-learning updates from Legilimens. This variant achieves higher accuracy than either system alone — showing up to 6.8% gains over Legilimens — particularly in scenes with frequent drift that outpaces the coverage of RECL's pre-trained models. However, this hybrid strategy incurs additional compute and storage overhead due to the need to maintain the full model zoo alongside a trainable base, leading to the takeaway that using Legilimens alone remains the more efficient option in serverly resource constrained settings.